\documentclass[conference]{IEEEtran}

\usepackage{amssymb}
\usepackage{graphicx}
\usepackage{bm}
\usepackage{amsmath,graphicx}
\usepackage{booktabs}
\usepackage{multirow}
\usepackage{verbatim}
\usepackage{booktabs}
\usepackage{algorithm}
\usepackage{algpseudocode}
\usepackage{amsfonts}
\usepackage{url}
\usepackage{enumerate}

\title{Unifying the Split Criteria of Decision Trees Using Tsallis Entropy}

\author{
\IEEEauthorblockN{Yisen Wang\IEEEauthorrefmark{1},  Chaobing Song\IEEEauthorrefmark{1}, Shu-Tao Xia\IEEEauthorrefmark{1}}

\IEEEauthorblockA{\IEEEauthorrefmark{1}Department of Computer Science and Technology, Tsinghua University, Beijing, China}
\IEEEauthorblockA{ Email: wangys14@mails.tsinghua.edu.cn, songchaobin@126.com, xiast@sz.tsinghua.edu.cn}
}

\begin{document}
\maketitle
\begin{abstract}
The construction of efficient and effective decision trees remains a key topic in machine learning because of their simplicity and flexibility. A lot of heuristic algorithms have been proposed to construct near-optimal decision trees. ID3, C4.5 and CART are classical decision tree algorithms and the split criteria they used are Shannon entropy, Gain Ratio and Gini index respectively. All the split criteria seem to be independent, actually, they can be unified in a Tsallis entropy framework. Tsallis entropy is a generalization of Shannon entropy and provides a new approach to enhance decision trees' performance with an adjustable parameter $q$. In this paper, a Tsallis Entropy Criterion (TEC) algorithm is proposed to unify Shannon entropy, Gain Ratio and Gini index, which generalizes the split criteria of decision trees. More importantly, we reveal the relations between Tsallis entropy with different $q$ and other split criteria. Experimental results on UCI data sets indicate that the TEC algorithm achieves statistically significant improvement over the classical algorithms.
\end{abstract}

% \begin{keyword}
% \MSC 41A05\sep 41A10\sep 65D05\sep 65D17
% \KWD Decision trees\sep Split criteria \sep Tsallis entropy

% %% MSC codes here, in the form: \MSC code \sep code
% %% or \MSC[2008] code \sep code (2000 is the default)
% \end{keyword}

% \end{frontmatter}

%\linenumbers

\section{Introduction}
\label{sec1}
The decision tree is a non-parametric supervised learning method used for classification and regression. Although the decision tree methods have been one of the first machine learning approaches, it remains an actively researched domain in machine learning. It is not only simple to understand and interpret, but also offers relatively good results, computational efficiency and flexibility. The general idea of decision trees is to predict unknown input instances by learning simple decision rules inferred from several known training instances. Decision trees are most often induced in the following top-down manner. A given data set is partitioned into a left and right subset by a split criterion test on attributes. The highest scoring partition which reduces the average uncertainty mostly is selected and the data set is partitioned accordingly into two child nodes, growing the tree by making the node be the parent of the two newly created child nodes. This procedure is applied recursively until some stopping conditions, e.g. maximum tree depth or minimum leaf size, are reached.

Generally speaking, split criterion is a fundamental issue in decision trees induction.
A large number of decision tree induction algorithms with different split criteria have been proposed. For example, the Iterative Dichotomiser 3 (ID3) algorithm is based on Shannon entropy \cite{quinlan1986induction}; the C4.5 algorithm is based on Gain Ratio which is considered as a normalized Shannon entropy \cite{quinlan2014c4}; while the Classification And Regression Tree (CART) algorithm is based on Gini index \cite{breiman1984classification}. These algorithms seem to be independent, and it is hard to judge which algorithm always outperforms others. Actually, it reflects one drawback of this kind of split criteria is that they lack adaptability to data sets. Numerous alternatives have been proposed for the adaptive entropy estimate \cite{nowozin2012improved,serrurier2015entropy}, but their statistical entropy estimates are too complex to lose the simplicity and comprehensibility of decision trees. Most of all, to the best of our knowledge, there is not a unified framework combining all the above criteria together. In addition, a series of papers have analyzed the importance of the split criterion \cite{buntine1992further,liu1994importance}. They demonstrated that different split criteria have substantial influence on the generalization error of the induced decision trees. This is the inspiration of our proposed new split criterion unifying and generalizing the classical split criteria.

To address the above issue, we propose a Tsallis entropy framework in this paper.
Tsallis entropy is a generalization of Shannon entropy with an adjustable parameter $q$ and is first introduced into decision trees in the prior work \cite{maszczyk2008comparison}. \cite{maszczyk2008comparison} only tested the performance of Tsallis entropy in C4.5 with some given $q$, but the relation between Tsallis entropy and other split criteria was not explored. And the unified framework was also not presented.
In this paper, we propose a Tsallis entropy based decision tree induction algorithm called TEC algorithm and analyze the correspondence between Tsallis entropy with different $q$ and other split criteria. Shannon entropy and Gini index are just two specific cases of Tsallis entropy with $q=1$ and $2$, while Gain Ratio is also can be considered as a normalized Tsallis entropy with $q = 1$. And Tsallis entropy indeed provides a new approach to improve the performance of decision trees with a tunable $q$ in a unified framework.
Experimental results on UCI data sets indicate that the TEC algorithm achieves statistically significant improvement over the classical algorithms without losing the strengths of decision trees.

The rest of this paper is organized as follows. Section 2 presents the background of Tsallis entropy. Section 3 outlines our proposed TEC algorithm. Section 4 exhibits experimental results. Section 5 summaries the work.

\section{Tsallis entropy}
Entropy is the measure of disorder in physical systems, or the measure of the amount of information that may be needed to specify the full microstates of the system \cite{frigg2011entropy}. In 1948, Shannon adopted entropy to information theory, called Shannon entropy \cite{shannon2001mathematical}, which is a measure of the uncertainty associated with a random variable.
\begin{equation}
H(X) = - \sum_{i =1}^n p(x_i) \ln p(x_i),
\end{equation}
where $X$ is a random variable that can take values $\{x_1,..., x_n\}$ and $p(x_i)$ is the corresponding probabilities of $x_i$. Shannon entropy is concave and attains maximum when $p(x_i) = 1/n, i = 1, 2, ..., n$.

There are two typical distributions observed in the macroscopic world, exponential distribution family and power-law heavy-tailed distribution family. However, we cannot characterize power-law heavy-tailed distribution through maximizing Shannon entropy subject to normal mean and variance. The reason is that Shannon entropy  implicitly assumes certain trade-off between contributions from the tails and the main mass of distribution \cite{maszczyk2008comparison}. It should be worthwhile to control this trade-off explicitly to characterize the two distribution family. Entropy measures that depend on powers of probability, $\sum_{i =1}^n p(x_i)^q$, can provide such control. Thus, some parameterized entropies have been proposed. A well-known generalization of this concept is Tsallis entropy \cite{tsallis1988possible}, which extends its applications to so-called non-extensive systems using an adjustable parameter $q$. Tsallis entropy can explain some physical systems that have complex behaviours such as long-range and long-memory interactions \cite{tsallis2009introduction}.

Tsallis entropy is defined by:
\begin{equation}
S_q(X) = \frac{1}{1-q}(\sum_{i=1}^n p(x_i)^q -1 ), \quad q \in \mathbb{R},
\end{equation}
which converges to Shannon entropy in the limit $q\to 1$,

\begin{align}
\lim_{q\to1} S_q(X) &= \lim_{q\to1} \frac{1}{1-q}(\sum_{i=1}^n p(x_i)^q -1 ) \nonumber \\
                    &= - \sum_{i =1}^n p(x_i) \ln p(x_i) \nonumber \\
                    &= H(X).
\end{align}

The relation to Shannon entropy can be made clearer by rewriting the definition in the form:
\begin{equation}
S_q(X) = - \sum_{i =1}^n p(x_i)^q \ln_q p(x_i),
\end{equation}
where
\begin{equation}
\ln_q(x) = \frac{x^{1-q}-1}{1-q}, \quad q \ne 1, x \ge 0
\end{equation}
is called the $q$-logarithmic function. And when $q\to 1$, $\ln_q(x)\to \ln(x)$.

Just like the exponential function to the logarithmic function, there is also the corresponding $q$-exponential function to $q$-logarithmic function.
\begin{equation}
e_q^x =
\begin{cases}
e^x & q = 1 \\
[1+(1-q)x]^{1/(1-q)} & q \ne 1 , 1+(1-q)x \ge 0 \\
0 & \text{otherwise}
\end{cases}.
\end{equation}

For $q < 0$, Tsallis entropy is convex. For $q =0$, Tsallis entropy is non-convex and non-concave. While for $q >0$, Tsallis entropy is concave, satisfying similar properties to Shannon entropy \cite{tsallis2009generalizing}. For instance, for $q >0$, $S_q \ge 0$, and $S_q$ is maximal at the uniform distribution.

Additivity is a crucial difference of the fundamental property between Shannon entropy and Tsallis entropy. For two independent random variables $X$ and $Y$, Shannon entropy has the additivity property:
\begin{equation}
H(X, Y) = H(X) + H(Y),
\end{equation}
however, Tsallis entropy $S_q(X)$ $(q\ne1)$ has the pseudo-additivity (also called $q$-additivity) property:
\begin{equation}
S_q(X, Y) = S_q(X) + S_q(Y) + (1-q)S_q(X)S_q(Y).
\end{equation}

Besides, Tsallis conditional entropy, Tsallis joint entropy and Tsallis mutual information are also derived similarly to Shannon entropy.
For the conditional probability $p(x|y) = p(X = x|Y =y)$ and the joint probability $p(x,y) = p(X=x,Y=y)$, Tsallis conditional entropy and Tsallis joint entropy \cite{abe2001nonadditive} are denoted by:
\begin{align} \label{Tc}
S_q(X|Y) &= - \sum_{x,y} p(x, y)^q \ln_q p(x|y), (q \ne 1)\\
S_q(X,Y) &= - \sum_{x,y} p(x, y)^q \ln_q p(x, y), (q \ne 1).
\end{align}
It is remarkable that Eq.(\ref{Tc}) can be easily deformed by
\begin{equation}
S_q(X|Y) = \sum_y p(y)^q S_q(X|y).
\end{equation}
The relation between the conditional entropy and the joint entropy is given by:
\begin{equation}\label{Tj}
S_q(X,Y) = S_q(X) + S_q(Y|X).
\end{equation}

Tsallis mutual information \cite{yamano2001information} is denoted as the difference between Tsallis entropy and Tsallis conditional entropy:
\begin{equation}\label{Mi}
I_q(X;Y) = S_q(X) - S_q(X|Y),
\end{equation}
and the chain rule of Tsallis mutual information for random variables $X_1, ..., X_n$ and $Y$ holds:
\begin{equation}\label{Cr}
I_q(X_1, ..., X_n; Y) = \sum_{i = 1}^n I_q(X_i; Y|X_1, ...., X_{i-1}).
\end{equation}
The relation among the conditional entropy, joint entropy and mutual information can be derived from Eq.(\ref{Tj}) and Eq.(\ref{Mi}):
\begin{equation}\label{dq}
S_q(Y|X) + S_q(X|Y) = S_q(X,Y) - I_q(X;Y).
\end{equation}

In summary, Tsallis entropy generalizes Shannon entropy with an adjustable parameter $q$ and has a wider range of applications.

\section{Tsallis Entropy Criterion (TEC) algorithm}
\label{Te}
One key issue in the procedure of decision tree induction is the split criterion. At every step, the decision tree chooses one pair of attribute and cutting point which makes the maximal impurity decrease to split the data and grow the tree. Therefore, the attribute chosen to split significantly affects the construction of decision trees and further influences the classification performance.

\subsection{Tree construction}

Given a data set $D_n = \{(X_i,Y_i)\}_{i = 1}^n$, $X_i \in \mathbb{R}^D$ with attributes $A_j$ $(j \in \{1,2,\dots,D\})$, and class label $Y_i \in \{1,2,\dots,k\}$. For each tree node, we search for every possible pair of attribute and cutting point to choose the optimal attribute and cutting point as follows: for a attribute $A_j$,
\begin{align}
\label{gain}
I(C_j) = T(D) - \frac{|D'|}{|D|}T(D') - \frac{|D''|}{|D|}T(D'')
\end{align}
Here $C_j$ is the candidate cutting point for attribute $A_j$, $D$ is the data set belonging to one node to be partitioned, and $D'$, $D''$ are the two child nodes that would be created if $D$ is partitioned at $C_j$. The function $T(D)$ is the impurity criterion, e.g. Tsallis entropy, which computes over the labels of the data which fall in the node. The pair of attribute $A_j$ and cutting point $C_j$ is chosen to construct the tree which maximizes $I(C_j)$.

The above procedure is applied recursively until some stopping conditions are reached. The stopping conditions consist of three principles: (i) The classification is achieved in a subset. (ii) No attributes are left for selection. (iii) The cardinality of a subset is not greater than the predefined threshold.

\subsection{Prediction}
Once the tree has been trained by the data as a classifier $g_n$, it can be used to predict for new unlabeled instances.

Decision tree makes prediction in a majority vote manner. For each class $k$,
\begin{equation}
\label{p1}
\eta^k (x) = \frac{1}{N(A_n(x))} \sum_{(X_i,Y_i) \in A_n(x)} \mathbb{I}(Y_i = k)
\end{equation}
where $A_n(x)$ denotes the leaf containing $x$, and $N(A_n(x))$ denotes the number of instances that located in $A_n(x)$. Then the tree prediction is the class that maximizes this value:
\begin{equation}
\label{p2}
g_n(x) = \arg \max_k \{\eta^k (x) \}
\end{equation}

\subsection{TEC algorithm}
Here, we summary our proposed Tsallis Entropy Criterion (TEC) algorithm in a pseudo-code format in Algorithm \ref{TEC}. Compared with the classical decision tree induction algorithms, the only difference is the split criterion. We use Tsallis entropy to replace the classical split criteria, e.g. Shannon entropy, Gain Ratio and Gini index. Actually, in the following subsection, we will see that Tsallis entropy unifies Shannon entropy, Gain Ratio and Gini index with different values of $q$.

\begin{algorithm}
\caption{TEC algorithm}
\label{TEC}
\begin{algorithmic}[1]
\State \textbf{Input}: Data $D_n$, Attributes $A$, Class $Y$
\State \textbf{Output}: \text{A decision tree}
\While{not satisfying stop condition}
    \For {each attribute $A_j$}
      \State $S \leftarrow domain(A_j)$
     \State // $S$ is the candidate cutting point set of attribute $A_j$
     \State // $C_j$ is one cutting point in the set $S$
    \For {\text{each} $C_j\in S$}
        \State $D' \leftarrow \{ d \in D_n | A_j(d) \le C_j \}$
        \State $D'' \leftarrow \{ d \in D_n | A_j(d) > C_j \}$
        \State // $d$ is one instance in Data $D_n$
        \State // $D'$, $D''$ are the two child data sets
        \State Compute $I(C_j)$ according to (\ref{gain})
    \EndFor
    \EndFor
    \State $C_{best} = \arg \max I(C_j)$
    \State $A_{best} \leftarrow A_j$
    \State // $A_{best}, C_{best}$ is the best pair of split attribute and cutting point
    \State Grow the tree using $A_{best}, C_{best}$
    \State Go to line 3 for $D'$ and $D''$
    \State // Recursively repeat the procedure
\EndWhile
\State \textbf{Return} Decision tree
\State // Tree is built by Nodes from the root to the leaf
\end{algorithmic}
\end{algorithm}

\subsection{Relations to other criteria}
As described above, Tsallis entropy unifies Shannon entropy, Gain Ratio and Gini index in a framework. In the following, we will reveal the relations between Tsallis entropy to other split criteria.

Tsallis entropy converges to Shannon entropy for $q\to1$:
\begin{align}
\lim_{q\to1} S_q(X) &= \lim_{q\to1} \frac{1}{1-q}(\sum_{i=1}^n p(x_i)^q -1 ) \nonumber \\
                    &= - \sum_{i =1}^n p(x_i) \ln p(x_i) \nonumber \\
                    &= H(X).
\end{align}
Besides, Gini index is exactly a specific case of Tsallis entropy with $q=2$:
\begin{align}
S_q(X)_{q = 2} &= \underbrace{\frac{1}{1-q}(\sum_{i=1}^n p(x_i)^q -1 )}_{q = 2} \nonumber \\
               &= 1- \sum_{i=1}^n p(x_i)^2 \nonumber \\
               &= \text{Gini index}
\end{align}

As for the Gain Ratio which adds a normalized factor compared with Information Gain, it can be seen as the normalized Information Gain. According to the Eq.(\ref{gain}), we can obtain:
\begin{align}
    \text{Gain Ratio} = \frac{\overbrace{H(D) - \frac{|D'|}{|D|}H(D') - \frac{|D''|}{|D|}H(D'')}^{\text{Information Gain}}}{H(\frac{|D'|}{|D|}, \frac{|D''|}{|D|})}
\end{align}
where $H$ represents Shannon entropy. If $H$ is replaced by Tsallis entropy, Gain Ratio is generalized to Tsallis Gain Ratio. Thus, Gain Ratio is also covered by the Tsallis entropy adding a normalized factor (Tsallis Gain Ratio) with $q = 1$.

In summary, Tsallis entropy unifies three kinds of split criteria, e.g. Shannon entropy, Gain Ratio and Gini index, and generalizes the split criterion of decision trees. As far as we know, this is the first time to unify common split criteria into a parametric framework. This is also the first time to reveal the correspondence between Tsallis entropy with different $q$ and other split criteria. The optimal $q$ for Tsallis entropy is obtained by cross-validation, which is usually not equal to $1$ or $2$. This implies better performance than the traditional split criteria. Although the optimal $q$ may be different for different data sets, it is associated with the properties of data sets. That is to say, the parameter $q$ enables the TEC algorithm to have adaptability and flexibility. Tsallis entropy indeed provides a new approach to improve decision trees' performance with a tunable $q$ in a unified framework.
In the Experiments section, we will see that the TEC algorithm achieves higher accuracy than classical algorithms with an appropriate $q$.

\section{Experiments}
\label{sec:typestyle}
As illustrated in section \ref{Te}, the TEC algorithm is based on Tsallis entropy with an adjustable parameter $q$ which consists of Tsallis entropy and Tsallis Gain Ratio split criteria. Tsallis entropy split criterion degenerates to Shannon entropy and Gini index with $q = 1$ and $q = 2$, respectively. With respect to Gain Ratio, Tsallis Gain Ratio (the normalized Tsallis entropy) also degenerates to Gain Ratio with $q = 1$.

\subsection{Evaluation Metric}
In order to quantitatively compare trees obtained by different methods, we choose accuracy to evaluate the effectiveness of the tree and the total number of the tree nodes to measure the tree complexity.

\subsection{Data Set Description}
As shown in Table \ref{tab2}, the $11$ UCI data sets \cite{Lichman:2013} are adopted to evaluate the proposed approaches. These data sets consist of three types, namely numeric, categorical and mixed data sets. Also, these data sets include two kinds of classification problems, binary and multi-class classification.
\newcommand{\tabincell}[2]{\begin{tabular}{@{}#1@{}}#2\end{tabular}}
\begin{table}[htp]
\centering
\caption{Data sets from UCI}
\label{tab2}
\begin{tabular}{ccccc}
\toprule[1.4pt]
Data Set & Type &  \tabincell{c} {No. of \\ instance} & \tabincell{c}{No. of \\features} & \tabincell{c}{No. of \\class}\\\hline
Yeast & numeric & 1484 & 8 & 10 \\
Glass& numeric &214&10&7\\
Vehicle & numeric & 946 & 18 & 4 \\
Wine & numeric & 178 & 13 & 3 \\
Haberman & numeric & 306 & 3 & 2 \\
Car & categorical & 1728 & 6& 4\\
Scale & categorical & 625 & 4 & 3 \\
Hayes & categorical & 160 & 5 & 3 \\
Monks & categorial & 432 & 7 & 2 \\
Abalone & mixed & 4139 & 8 & 18 \\
Cmc & mixed & 1473 & 9 & 3 \\
\bottomrule[1.4pt]
\end{tabular}
\end{table}

\subsection{Experiment Setup}
The decision trees with different split criteria, e.g. Gain Ratio, Shannon entropy, Gini index, Tsallis entropy and Tsallis Gain Ratio, are implemented in Python. We refer to the CART algorithm implementation on scikit-learn platform \cite{scikit-learn} and the C4.5 algorithm implementation of J48 in Weka \cite{hall2009weka}. In each data set, we first partition the data into the training set and test set randomly where the test set holds $30\%$. Then in the training set, we do a grid search using 10-fold cross-validation to determine the the values of $q$ in Tsallis entropy and Tsallis Gain Ratio. Maybe the optimal $q$ for Tsallis entropy and Tsallis Gain Ratio are different, but for the fair comparison we choose the same $q$, e.g. optimal $q$ for Tsallis entropy. Besides, the minimal leaf size is set to $5$ to avoid overfitting. After the parameter selection, the above best parameters are fixed. Then, a decision tree is trained by the training data without post-pruning and evaluated by the test data. The procedure from the training-test data partition to the evaluation is repeated 10 times to reduce the influence of randomness.

\subsection{Results}
Figure 1 gives an intuitive exhibition of the influence of different values of parameter $q$ in Tsallis entropy for the Glass data set. Figure 1 (a) illustrates that the accuracy is sensitive to the change of $q$ and the highest accuracy is obtained at $q = 2.6$. Figure 1 (b) shows that the tree complexity has different responds to the change of $q$ as accuracy and the lowest tree complexity is achieved at $q = 3.9$. It should be noted that there are different strategies to choose $q$ for various purpose, e.g. highest accuracy or lowest complexity or trade-off, which is also a reflection of the TEC algorithm's adaptability for data sets. In this paper, we choose the highest accuracy principle for the choice of $q$.

Table \ref{tab3} reports the accuracy and complexity results of different criteria for different data sets. The highest accuracy and lowest complexity on each data set are in boldface. As expected, the performance of TEC outperforms ID3, CART and C4.5 due to the fact that Tsallis entropy is a generalization of Shannon entropy, Gini index and Gain Ratio. In respect to the two kinds of the TEC algorithm, e.g. Tsallis entropy and Tsallis Gain Ratio, no one can prevail another one absolutely. The results indicates that Tsallis entropy prefers high accuracy while Tsallis Gain Ratio prefers low complexity. The reason lies on the normalized factor which has influence in the tree structure to some extent. In addition, compared with Shannon entropy and Gini index, Tsallis entropy achieves better performance in accuracy and complexity. Tsallis Gain Ratio also obtains better results compared with Gain Ratio.  Three Wilcoxon signed ranked tests \cite{demvsar2006statistical} on accuracy (Tsallis entropy vs Shannon entropy, Tsallis entropy vs Gini index, Tsallis Gain Ratio vs Gain Ratio) all reject the null hypothesis of equal performance at a p-value less than $0.01$. The results show that the TEC algorithm with appropriate $q$ achieves a average $4\%$ statistically significant improvement in accuracy and maintains a lower complexity.

In terms of optimal value of $q$, we find a fuzzy trend from Table \ref{tab3} that the more of class number, the smaller $q$ value is tended, e.g. for numeric type data sets from Yeast to Haberman, $q$ is increasing while the class number is decreasing (exception for Vehicle). In this paper, we choose the optimal value of $q$ using cross-validation method, but we conjecture that the values of $q$ is associated with the properties of data sets. For example, the Car data set, all the algorithms presents almost the same results which reflects the data set is not sensitive to the parameter $q$. The relation between the $q$ and the properties of data sets will be discussed in the future work.

\begin{figure}[htb]
\begin{minipage}[b]{1.0\linewidth}
  \centering
  \centerline{\includegraphics[width=8.5cm]{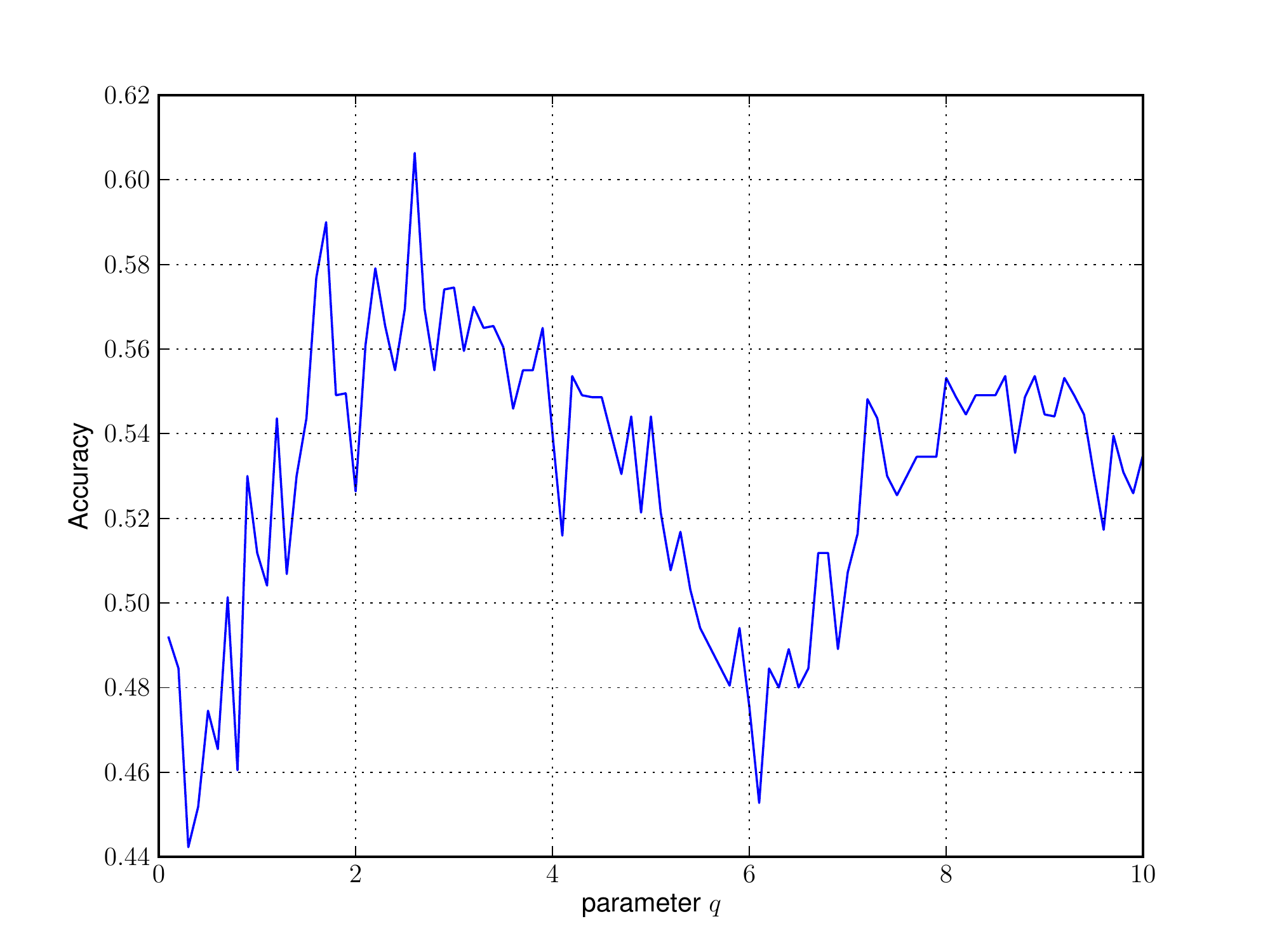}}
  \centerline{(a) Accuracy with different values of $q$}\medskip
\end{minipage}

\begin{minipage}[b]{1.0\linewidth}
  \centering
  \centerline{\includegraphics[width=8.5cm]{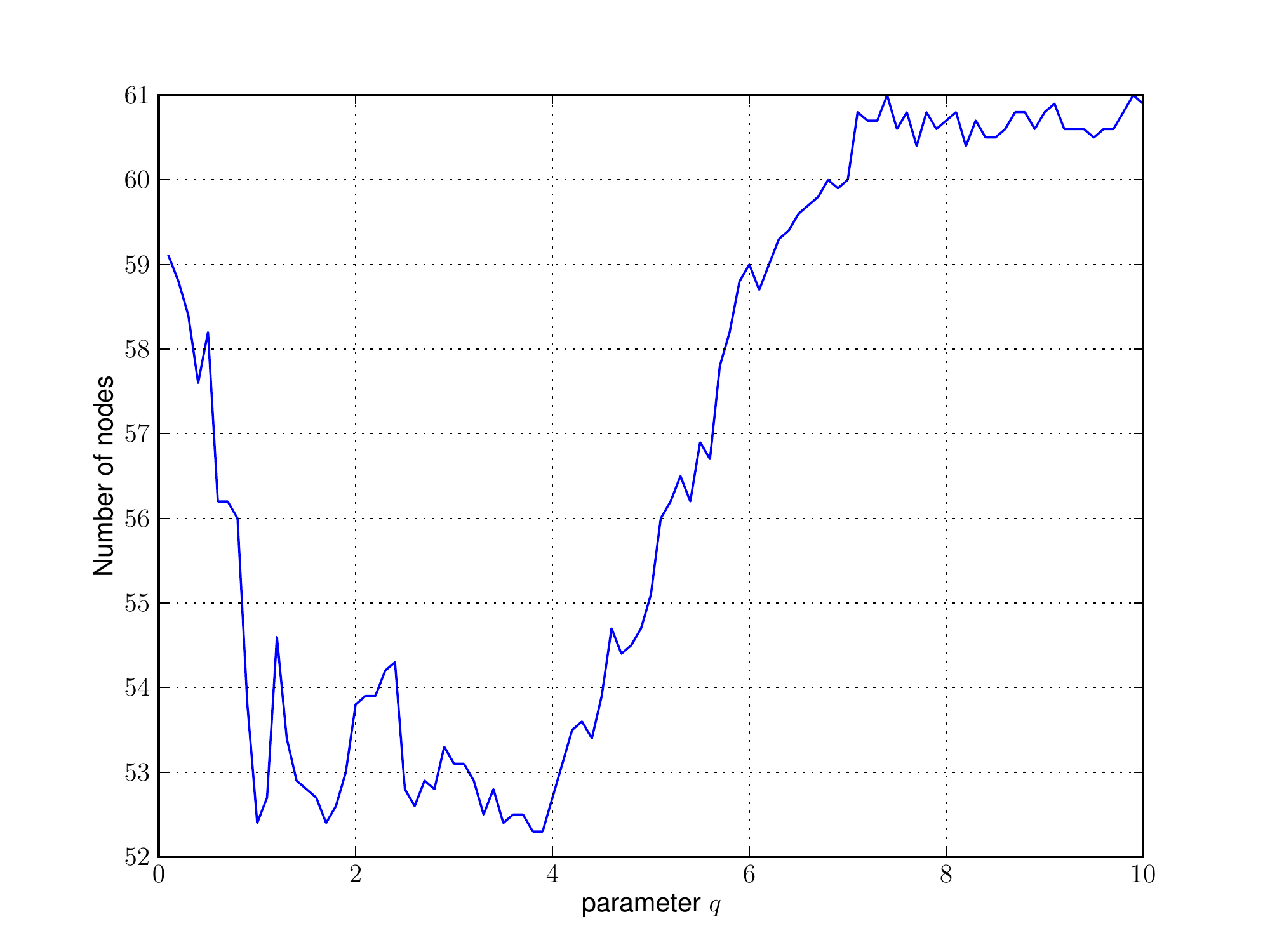}}
  \centerline{(b) Tree complexity with different values of $q$}\medskip
\end{minipage}
\caption{Influence of parameter $q$ in tree classification accuracy and tree complexity for Glass data set}
\label{fig:res}
\end{figure}

\begin{table*}[htbp]
\centering
\caption{Comparison of different decision tree split criteria}
\label{tab3}
\begin{tabular}{c|cc|cc|cc|ccc|cc}
\toprule[1.4pt]
\multirow{2}{*}{Data Set}  &\multicolumn{2}{c}{ \tabincell{c} {Shannon entropy \\ (ID3)}} & \multicolumn{2}{c}{ \tabincell{c} {Gini index \\ (CART)}} &  \multicolumn{2}{c}{ \tabincell{c} {Gain Ratio \\ (C4.5)}} & \multicolumn{3}{c}{ \tabincell{c} {Tsallis entropy \\ (TEC)}} & \multicolumn{2}{c}{ \tabincell{c} {Tsallis Gain Ratio \\ (TEC)}} \\\cline {2-12}
 & \tabincell{c} {Accuracy \\(\%)} &  \tabincell{c} {No. of \\nodes} & \tabincell{c} {Accuracy \\(\%)} &  \tabincell{c} {No. of \\nodes} & \tabincell{c} {Accuracy \\(\%)} & \tabincell{c} {No. of \\nodes} & \tabincell{c} {Accuracy \\(\%)} & \tabincell{c} {No. of \\nodes} & $q$ & \tabincell{c} {Accuracy \\(\%)} & \tabincell{c} {No. of \\nodes}\\\hline
Yeast & 52.8 & 199 & 51.8 & 196.6 & 52.1 & 326.2 & \bf{56.9} & \bf{195.8} & 1.4 & 51.2 & 197.1\\
Glass & 51.2 & 52.4 & 52.6 & 53.8 & 44.2 & 52 & \bf{60.6} & 52.6 & 2.6 & 53.1 & \bf{51.5} \\
Vehicle & 71.7 & 103 & 70.2 & \bf{100} & 72.3 & 147.2 & \bf{73.8} & 111.0 & 0.6 & 73.4 & 135.7 \\
Wine & 92.9 & 12.0 & 90.0 & 12.0 & 92.4 & 9.4 & \bf{95.9} & 9.6 & 3.1 & 92.9 & \bf{9.2}\\
Haberman & 70.3 & 32.2 & 70.3 & 33.0 & 72.8 & 33.0 & 74.2 & 33.2 & 7.1 & \bf{74.8} & \bf{33.0} \\
Car & 98.2 & 106.4 & 98.1 & 106.8 & \bf{98.5} & 106.5 & 98.3 & \bf{106.2} & 0.8 & 98.4 & 106.6 \\
Scale & 75.9 & 97.6 & 76.1 & 97.2 & 74.5 & 77.0 & 78.2 & 93.1 & 3.1 & \bf{78.5} & \bf{77.0} \\
Hayes & 81.5 & 28.8 & 80.0 & 25.3 & 79.2 & 19.6 & \bf{82.3} & 19.5 & 8.6 & 81.5 & \bf{19.2} \\
Monks & 51.9 & 89.0 & 52.1 & 88.6 & 52.9 & 88.0 & \bf{57.3} & 89.6 & 8.9 & 54.9 & \bf{88.0} \\
Abalone & 25.4 & 89.2 & 25.0 & 85.8 & 20.3 & \bf{84.3} & \bf{26.8} & 86.2 & 0.8 & 25.7 & 85.1 \\
Cmc & 49.1 & 267.0 & 47.4 & 264.0 & 45.7 & 242.8 & \bf{52.0} & 264.2 & 1.2 & 47.8 & \bf{242.1}\\
\bottomrule[1.4pt]
\end{tabular}
\end{table*}

\section{Conclusions}
\label{con}
In this paper, we present and evaluate Tsallis entropy for enhancing decision trees in a fundamental issue, e.g. split criterion. We unify the classical split criteria into a parametric framework and propose the TEC algorithm with Tsallis entropy split criterion which generalizes Shannon entropy, Gain Ratio and Gini index through an adjustable parameter $q$.
Most of all, we reveal the relations between Tsallis entropy with different $q$ and other split criteria.
Experimental results indicate that, with appropriate $q$, the TEC algorithm achieves a average $4\%$ statistically significant improvement in accuracy.
Nevertheless, the approaches have limitations that need to be addressed in the future, such as, the estimate method for parameter $q$ in place of current cross-validation method. Furthermore, Tsallis entropy also has potential applications beyond decision trees, for instance, Random Forest and Bayesian network, to be investigated in future work.

% Thus, Tsallis entropy framework indeed provides a new approach to enhance the decision trees' performance without hurting their simplicity and comprehensibility.

\section*{Acknowledgments}
This research is supported in part by the 973 Program of China (No. 2012CB315803), the National Natural Science Foundation of China (No. 61371078, 61375054), and the Research Fund for the Doctoral Program of Higher Education of China (No. 20130002110051).

\bibliographystyle{IEEEtran}
\bibliography{refs}

\begin{thebibliography}{10}
\providecommand{\url}[1]{#1}
\csname url@samestyle\endcsname
\providecommand{\newblock}{\relax}
\providecommand{\bibinfo}[2]{#2}
\providecommand{\BIBentrySTDinterwordspacing}{\spaceskip=0pt\relax}
\providecommand{\BIBentryALTinterwordstretchfactor}{4}
\providecommand{\BIBentryALTinterwordspacing}{\spaceskip=\fontdimen2\font plus
\BIBentryALTinterwordstretchfactor\fontdimen3\font minus
  \fontdimen4\font\relax}
\providecommand{\BIBforeignlanguage}[2]{{%
\expandafter\ifx\csname l@#1\endcsname\relax
\typeout{** WARNING: IEEEtran.bst: No hyphenation pattern has been}%
\typeout{** loaded for the language `#1'. Using the pattern for}%
\typeout{** the default language instead.}%
\else
\language=\csname l@#1\endcsname
\fi
#2}}
\providecommand{\BIBdecl}{\relax}
\BIBdecl

\bibitem{quinlan1986induction}
J.~R. Quinlan, ``Induction of decision trees,'' \emph{Machine Learning},
  vol.~1, no.~1, pp. 81--106, 1986.

\bibitem{quinlan2014c4}
J.~R. Quinlan, \emph{C4. 5: programs for machine learning}.\hskip 1em plus
  0.5em minus 0.4em\relax Morgan Kaufmann Publishers, 1993.

\bibitem{breiman1984classification}
L.~Breiman, J.~Friedman, C.~J. Stone, and R.~A. Olshen, \emph{Classification
  and regression trees}.\hskip 1em plus 0.5em minus 0.4em\relax CRC press,
  1984.

\bibitem{nowozin2012improved}
S.~Nowozin, ``Improved information gain estimates for decision tree
  induction,'' in \emph{Proceedings of the 29th International Conference on
  Machine Learning (ICML-12)}.\hskip 1em plus 0.5em minus 0.4em\relax ACM,
  2012, pp. 297--304.

\bibitem{serrurier2015entropy}
M.~Serrurier and H.~Prade, ``Entropy evaluation based on confidence intervals
  of frequency estimates: Application to the learning of decision trees,'' in
  \emph{Proceedings of the 32nd International Conference on Machine Learning
  (ICML-15)}.\hskip 1em plus 0.5em minus 0.4em\relax ACM, 2015, pp. 1576--1584.

\bibitem{buntine1992further}
W.~Buntine and T.~Niblett, ``A further comparison of splitting rules for
  decision-tree induction,'' \emph{Machine Learning}, vol.~8, no.~1, pp.
  75--85, 1992.

\bibitem{liu1994importance}
W.~Z. Liu and A.~P. White, ``The importance of attribute selection measures in
  decision tree induction,'' \emph{Machine Learning}, vol.~15, no.~1, pp.
  25--41, 1994.

\bibitem{maszczyk2008comparison}
T.~Maszczyk and W.~Duch, ``Comparison of shannon, renyi and tsallis entropy
  used in decision trees,'' in \emph{Proceedings of the 17th International
  Conference on Artificial Intelligence and Soft Computing (ICAISC-08)}.\hskip
  1em plus 0.5em minus 0.4em\relax Springer, 2008, pp. 643--651.

\bibitem{frigg2011entropy}
R.~Frigg and C.~Werndl, ``Entropy-a guide for the perplexed,''
  \emph{Probabilities in physics}, 2011.

\bibitem{shannon2001mathematical}
C.~Shannon, ``A mathematical theory of communication,'' \emph{Bell System
  Technical Journal}, vol.~27, no.~3, pp. 379--423, 1948.

\bibitem{tsallis1988possible}
C.~Tsallis, ``Possible generalization of boltzmann-gibbs statistics,''
  \emph{Journal of Statistical Physics}, vol.~52, no. 1-2, pp. 479--487, 1988.

\bibitem{tsallis2009introduction}
C.~Tsallis, \emph{Introduction to nonextensive statistical mechanics}.\hskip
  1em plus 0.5em minus 0.4em\relax Springer, 2009.

\bibitem{tsallis2009generalizing}
C.~Tsallis, ``Generalizing what we learnt: Nonextensive statistical
  mechanics,'' in \emph{Introduction to Nonextensive Statistical
  Mechanics}.\hskip 1em plus 0.5em minus 0.4em\relax Springer, 2009, pp.
  37--106.

\bibitem{abe2001nonadditive}
S.~Abe and A.~Rajagopal, ``Nonadditive conditional entropy and its significance
  for local realism,'' \emph{Physica A: Statistical Mechanics and its
  Applications}, vol. 289, no.~1, pp. 157--164, 2001.

\bibitem{yamano2001information}
T.~Yamano, ``Information theory based on nonadditive information content,''
  \emph{Physical Review E}, vol.~63, no.~4, p. 046105, 2001.

\bibitem{Lichman:2013}
M.~Lichman, ``{UCI} machine learning repository,''
  \url{http://archive.ics.uci.edu/ml}, 2013.

\bibitem{scikit-learn}
F.~Pedregosa, G.~Varoquaux, A.~Gramfort, V.~Michel, B.~Thirion, O.~Grisel,
  M.~Blondel, P.~Prettenhofer, R.~Weiss, V.~Dubourg, J.~Vanderplas, A.~Passos,
  D.~Cournapeau, M.~Brucher, M.~Perrot, and E.~Duchesnay, ``Scikit-learn:
  Machine learning in {P}ython,'' \emph{Journal of Machine Learning Research},
  vol.~12, pp. 2825--2830, 2011.

\bibitem{hall2009weka}
M.~Hall, E.~Frank, G.~Holmes, B.~Pfahringer, P.~Reutemann, and I.~H. Witten,
  ``The weka data mining software: an update,'' \emph{ACM SIGKDD explorations
  newsletter}, vol.~11, no.~1, pp. 10--18, 2009.

\bibitem{demvsar2006statistical}
J.~Dem{\v{s}}ar, ``Statistical comparisons of classifiers over multiple data
  sets,'' \emph{The Journal of Machine Learning Research}, vol.~7, pp. 1--30,
  2006.

\end{thebibliography}

% \section*{Supplementary Material}

% Supplementary material that may be helpful in the review process should
% be prepared and provided as a separate electronic file. That file can
% then be transformed into PDF format and submitted along with the
% manuscript and graphic files to the appropriate editorial office.

\end{document}